\documentclass[lettersize,journal]{IEEEtran}
\usepackage{amsmath,amsfonts}
\usepackage{algorithmic}
\usepackage{algorithm}
\usepackage{array}
\usepackage[caption=false,font=normalsize,labelfont=sf,textfont=sf]{subfig}
\usepackage{textcomp}
\usepackage{stfloats}
\usepackage{url}
\usepackage{verbatim}
\usepackage{graphicx}
\usepackage{cite}
\usepackage{xcolor}
\usepackage{multirow}
\usepackage{hyperref}
\begin{document}

\title{Joint Person Identity, Gender and Age Estimation from Hand Images using Deep Multi-Task Representation Learning}

%\author{IEEE Publication Technology,~\IEEEmembership{Staff,~IEEE,}
%        % <-this % stops a space
%\thanks{This paper was produced by the IEEE Publication Technology Group. They are in Piscataway, NJ.}% <-this % stops a space
%\thanks{Manuscript received April 19, 2021; revised August 16, 2021.}}

\author{Nathanael L. Baisa
%\author{Nathanael L. Baisa, Bryan Williams, Hossein Rahmani, Plamen Angelov, Sue Black  

\thanks{Nathanael L. Baisa is with the School of Computer Science and Informatics, De Montfort University, Leicester LE1 9BH, UK. Email: nathanael.baisa@dmu.ac.uk.}
%\thanks{Bryan Williams, Hossein Rahmani, Plamen Angelov and Sue Black are with the School of Computing and Communications, Lancaster University, Lancaster LA1  4WA, UK. Email: \{b.williams6, h.rahmani, p.angelov, sue.black\}@lancaster.ac.uk.}
}

%% The paper headers


\maketitle

\begin{abstract}
In this paper, we propose a multi-task representation learning framework to jointly estimate the identity, gender and age of individuals from their hand images for the purpose of criminal investigations since the hand images are often the only available information in cases of serious crime such as sexual abuse. We investigate different up-to-date deep learning architectures and compare their performance for joint estimation of identity, gender and age from hand images of perpetrators of serious crime. To simplify the age prediction, we create age groups for the age estimation. We make extensive evaluations and comparisons of both convolution-based and transformer-based deep learning architectures on a publicly available 11k hands dataset. Our experimental analysis shows that it is possible to efficiently estimate not only identity but also other attributes such as gender and age of suspects jointly from hand images for criminal investigations, which is crucial in assisting international police forces in the court to identify and convict abusers. The source code is available at \textcolor{red}{\url{https://github.com/nathanlem1/IGAE-Net}}. 

\end{abstract}


\section{Introduction} \label{sec:intro}

Recognition of individuals using body parts or behavioural characteristics, often termed as biometric identitification~\cite{SheAmiHan19, JaiDebEng22}, has recently attracted remarkable attention for numerous applications. The primary biometric traits, also called biometric modalities, used for the identification of an individual include face~\cite{DenGuoZaf19}, body~\cite{Nat22}, hand~\cite{Mah19}, fingerprint~\cite{EngCaoJai19}, iris~\cite{ZhaKum17}, gait~\cite{MurShiMak15} and voice~\cite{ChuNagZis18}. Personal attributes or ancillary information such as gender, age, (hair, skin, eye) color, ethnicity, height, weight, etc. which are also referred to as soft biometrics can be inexpensively obtained from the primary biometric traits~\cite{DanEliRos16}. The recognition of an individual can be improved by fusing different biometric modalities or by fusing biometric modalities with soft biometrics~\cite{SinSinRos19}. Hand images are often the only available information in cases of serious crime such as sexual abuse. This is because abusers usually hide their identity by covering, for instance, their faces; however, their hands often remain visible. Hand images also deliver discriminative features for biometric person recognition. They not only have less variability when compared to other biometric modalities but also have strong and diverse features which remain relatively stable after adulthood~\cite{BaiWilHosGPA21, YimChaShu20, AttAkhCha21}. Because of this, there is a strong potential to investigate hand images captured by digital cameras for persons recognition (preferably with their gender and age predictions), especially for criminal investigations in uncontrolled environments, which is crucial in assisting international police forces in the court to identify and convict abusers.

Full hand images have been used in~\cite{Mah19} to identify a person using both traditional and deep learning methods. Similar appraoch has also been used in~\cite{YimChaShu20} with additional data type for fusion, near-infrared (NIR) images, in addition to the RGB hand images. However, these methods are not an end-to-end. An end-to-end approaches considering both horizontal and vertical uniform partitioning~\cite{BaiWilHosGPA21} and multi-branch network with attention mechanism~\cite{BaiWilHosMBA21} have been proposed using full hand images as input. Knuckle patterns of dorsal hand images have been used in~\cite{KumXu16, AttAkhCha21, RitHosRic21} for person identification. The work in~\cite{MonPlaBry21} has combined fingernails and dorsal knuckle patterns for identifying individuals. Full hand images have been combined with knuckle patterns in~\cite{ZahSeyRam22} for person recognition. Age estimation from hand images is also proposed in~\cite{AbdGueAbd20}. Facial images-based age and gender prediction methods are proposed in~\cite{AmiMehElh20, GarRaySin21}; also with additional ethnicity prediction~\cite{HanJai14} or identity recognition~\cite{Savchenko19}. However, all these hand-based methods recognize only the identity of an individual or only age of an individual from hand images. Unlike these methods, our proposed deep multi-task representation learning framework jointly estimates the identity, gender and age of individuals from their hand images for the purpose of criminal investigations.

In this work, we propose a multi-task representation learning framework to jointly estimate the identity, gender and age of individuals from their hand images for the purpose of criminal investigations. We investigate the performance of different deep learning architectures and compare their performance for joint estimation of identity, gender and age from hand images of perpetrators of serious crime. To simplify the age prediction, we create age groups for the age estimation. Our contributions can be summarized as follows.

\begin{enumerate}
\item We propose a multi-task representation learning framework to jointly estimate the identity, gender and age of individuals from their hand images.
\item We investigate different up-to-date deep learning architectures and compare their performance for joint estimation of identity, gender and age from hand images. 
\item We make extensive evaluations and comparisons on a publicly available 11k~\cite{Mah19} hands dataset.
\end{enumerate}

The rest of the paper is organized as follows. The proposed method is described in Section~\ref{sec:proposedMethod} including the overall architecture and the loss function. The experimental results are analyzed and compared in Section~\ref{sec:experimentalResults} followed by the main conclusion along with suggestion for future work in Section~\ref{sec:Conclusion}.

\section{Proposed Method} \label{sec:proposedMethod}

In this section, we introduce the overall architecture of the identity, gender and age estimation network (IGAE-Net) and the used loss function.

\subsection{Network Architecture Overview}

The overall architecture of the proposed IGAE-Net is given in Fig.~\ref{fig:IGAE-Net}. We use different deep learning architectures as backbone network. We keep the original structure of each deep learning architecture used as backbone before its classifier layer remain the same when we modify each backbone network to produce the IGAE-Net. We create 3 independent branches just before the classifier layer of each backbone network to produce three heads: identity head, gender head and age head. Each head follows the original classifier layer structure of the backbone network. We use both convolution-based and transformer-based up-to-date deep learning architectures as backbone network for detailed performance comparisons.  

\textbf{CNN-based Backbone}: For the case of the convolution-based architectures, we use ResNet50~\cite{HeZhaSun16}, DenseNet121~\cite{HuaLiuVan17} and ConvNeXt-Tiny~\cite{ZhuHanCha22}. 

\textbf{Transformer-based Backbone}: For the case of the transformer-based architectures, we use ViT-B-16~\cite{AleLucAle21}, Swin-T~\cite{ZeYutYue21} and MaxViT-T~\cite{TuTalZha22}. 

Each of the identity head, gender head and age head, which acts as classification layer, is implemented independently using a fully-connected (FC) layer (with additional normalization layer(s) for the ConvNeXt-Tiny and MaxViT-T) followed by a softmax function to jointly predict the identity, gender and age group of each input hand image, respectively.

\subsection{Loss Function}

\begin{figure*}[t]%[htbp] %[t]%[!htb] %[t]%[!h]
\begin{center}
  \includegraphics[width=0.8\linewidth]{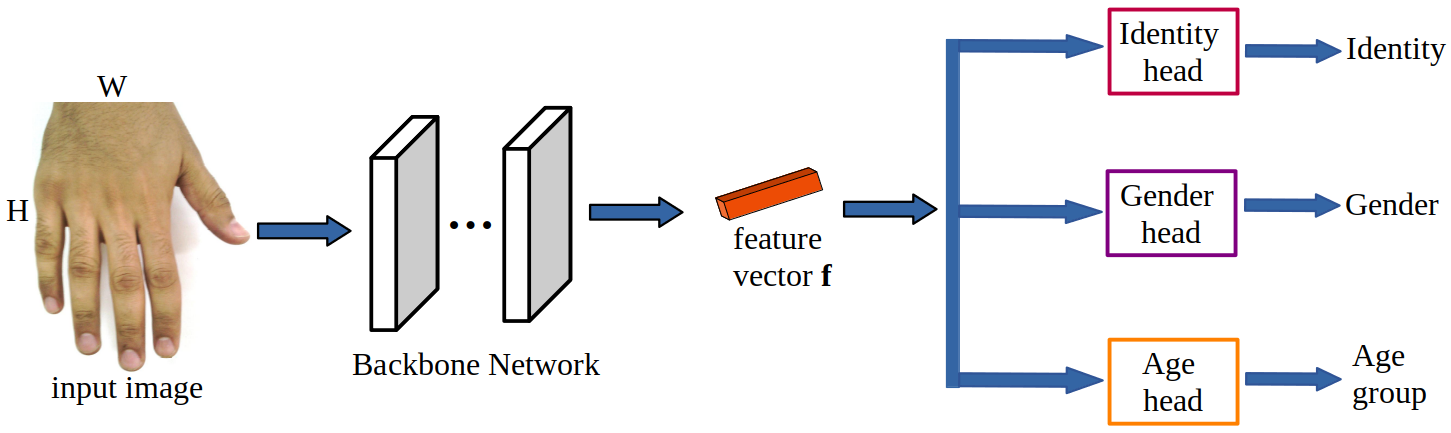} \\
\end{center}
   \caption{Structure of IGAE-Net. Given an input hand image, feature vector \textbf{f} is obtained by passing it through the backbone network and then is fed into each head (identity head, gender head and age head). The identity head, gender head and age head predict the identity, gender and age group of the input image, respectively.}
\label{fig:IGAE-Net}
\end{figure*}

The IGAE-Net is optimized during training by minimizing the loss function $\mathcal{L}$ consisting of the sum of cross-entropy losses over the predictions from the 3 heads i.e. the identity head, the gender head and the age head for multi-output classification, as given in Eq.~\ref{eqn:totalLoss}. Given the input hand image, classifier from each head predicts its corresponding output i.e. identity, gender and age group of the input image as shown in Fig.~\ref{fig:IGAE-Net}. 

\begin{equation}
    \mathcal{L} = \sum_{l=1}^3 \mathcal{L}_{l,xent} 
\label{eqn:totalLoss}
\end{equation}
\noindent For the learned features $\textbf{f}_i$ (for sample $i$), the cross-entropy loss (softmax loss) with label smoothing~\cite{SzeVanIof16} is given as: 

\begin{equation}
    \mathcal{L}_{l,xent} = -\frac{1}{N}\sum_{i=1}^N \sum_{j=1}^C q_{y_{i,j}} \log \frac{e^{\textbf{W}^T_{j} \textbf{f}_i + b_{j}}}{\sum_{c=1}^C e^{\textbf{W}^T_c \textbf{f}_i + b_c}} 
\label{eqn:xent}
\end{equation}
\noindent where N is the batch-size, C is the number of classes (identities) in the training dataset, and $\textbf{W}_c$ and $b_c$ are weight vector and bias for class $c$, respectively. Note that $z_c = \textbf{W}^T_c \textbf{f}_i + b_c$ are the logits or unnormalized probabilities. The ground-truth distribution over labels $q_{y_{i,j}}$ by including label smoothing can be given as

\begin{equation}
    q_{y_{i,j}} =
\begin{cases}
    1 - \frac{C-1}{C} \epsilon, & \text{if } y_{i,j} = y\\
    \frac{1}{C} \epsilon,              & \text{otherwise}
\end{cases}
\label{eq:labelSmoothing}
\end{equation}
\noindent where $y$ is ground-truth label (for sample $i$ and class $j$) and $\epsilon$ is a smoothing value.

\noindent

\section{Experiments}  \label{sec:experimentalResults}

\subsection{Dataset}

We use the 11k hands dataset\footnote{\url{https://sites.google.com/view/11khands}}~\cite{Mah19} which has 190 subjects (identities) with genders and varying ages between 18 - 75 years old. Any hand image containing accessories is excluded from the data to avoid any potential bias as in~\cite{BaiWilHosGPA21}. The dataset is then divided into right dorsal, left dorsal, right palmar and left palmar sub-datasets.  After excluding accessories and dividing the dataset into the sub-datasets,  the right dorsal has 143 identities, the left dorsal has 146, the right palmar has 143 and the left palmar has 151 identities. We take the samples in random order before dividing each sub-dataset into training and test sets equally based on the number of samples (in each class or subject). 

\begin{figure}[htbp]%[!htb] %[t]%[!h]
  \begin{center}
   \subfloat[] %[Channel Attention Module (CAM)]
  {\label{fig:AgeDistribution} \includegraphics[width=0.70\linewidth]{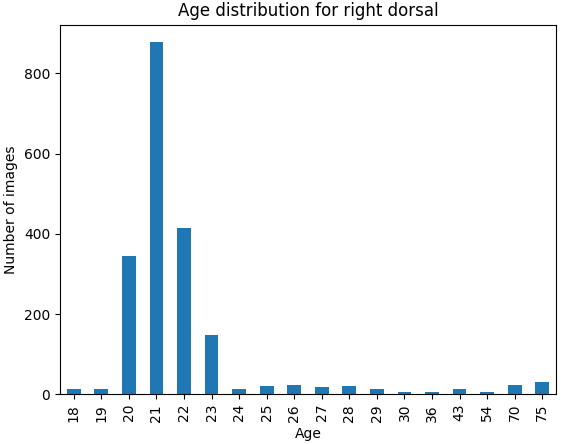}} \\%& %\\ %height=0.46
  \subfloat[] %[Spatial Attention Module with Relative Positional Encodings (SAM-RPE)]
  {\label{fig:AgeGroupDistribution} \includegraphics[width=0.70\linewidth]{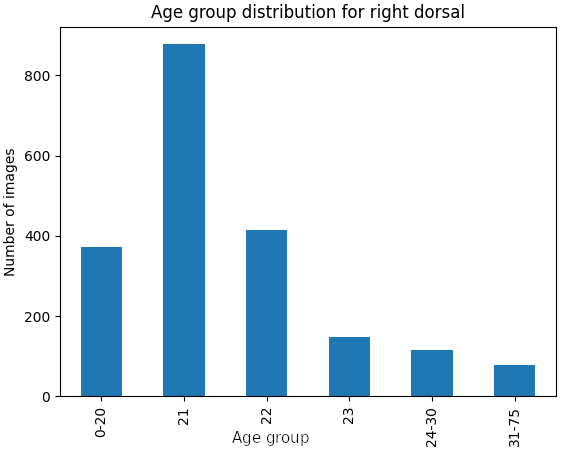}}  %& %\\ %height=0.18
  \end{center}
   \caption{Age statistics for right dorsal of the 11k hands dataset~\cite{Mah19}: (a) Age distribution, (b) Age group distribution. The number of images per age or age group is shown.}
  \label{fig:Age_DistributionBoth}
\end{figure}

The age distribution of the 11k hands dataset (right dorsal as an example) is shown in Fig.~\ref{fig:AgeDistribution}. As can be seen in this figure, the largest portion of the images belongs to individuals with age between 20 and 23. The number of images for persons with other ages is very low. This poses a problem to properly learn the ages of individuals, especially by treating the prediction task as a classification task. To alleviate this problem and simplify the age prediction, we group images of persons whose age lies in some range as shown in Fig.~\ref{fig:AgeGroupDistribution}. The grouping is done into 6 classes (age groups) as can be observed in Fig.~\ref{fig:AgeGroupDistribution}. In this manner, we can predict the age group of the persons instead of their exact age. Accordingly, the identity statistics, gender statistics and age groups statistics of each sub-dataset used in this paper are given in Table~\ref{tbl:HandDatasetsIdentity}, Table~\ref{tbl:HandDatasetsGender}  and Table~\ref{tbl:HandDatasetsAge}, respectively.

\noindent

\begin{table} [htbp]%[!htb]%[!h]%[tb]
\caption{\normalfont{Identity statistics of the 11k hands dataset~\cite{Mah19} for right dorsal (D-r), left dorsal (D-l), right palmar (P-r) and left palmar (P-l) sub-datasets. Number of identities (ids) and number of images (samples) are shown for each sub-dataset.}}
\label{tbl:HandDatasetsIdentity}
\begin{center}
  \begin{tabular}{|l|l|l|l|l|}
    \hline
    & D-r of 11k & D-l of 11k & P-r of 11k & P-l of 11k \\
    \hline
	\# ids & 143 & 146 & 143 & 151   \\ 
	\# images & 2004 & 1869 & 1965 & 2027   \\
	\hline 
  \end{tabular}
\end{center}
%\caption{Statistics of hand-based person Re-Id datasets used in this paper: 11k~\cite{Mah19} and HD~\cite{KumXu16}. Number of identities (ids), number of images and number of cameras are shown for train set, query set and gallery set of each dataset.}
%\label{tbl:HandDatasets}
\end{table}
\noindent

\begin{table} [htbp]%[!htb]%[!h]%[tb]
\caption{\normalfont{Gender statistics of the 11k hands dataset~\cite{Mah19} for right dorsal (D-r), left dorsal (D-l), right palmar (P-r) and left palmar (P-l) sub-datasets. Number of images (samples) for male and female are shown.}}
\label{tbl:HandDatasetsGender}
\begin{center}
  \begin{tabular}{|l|l|l|l|l|}
    \hline
    & D-r of 11k & D-l of 11k & P-r of 11k & P-l of 11k \\
    \hline
	Male & 909 & 846 & 948 & 949   \\ 
	Female & 1095 & 1023 & 1017 & 1078   \\
	\hline 
  \end{tabular}
\end{center}
\end{table}
\noindent

\begin{table} [htbp]%[!htb]%[!h]%[tb]
\caption{\normalfont{Age groups statistics of the 11k hands dataset~\cite{Mah19} for right dorsal (D-r), left dorsal (D-l), right palmar (P-r) and left palmar (P-l) sub-datasets. Number of images (samples) for each age group is shown.}}
\label{tbl:HandDatasetsAge}
\begin{center}
  \begin{tabular}{|l|l|l|l|l|}
    \hline
    & D-r of 11k & D-l of 11k & P-r of 11k & P-l of 11k \\
    \hline
	0-20  & 372 & 328 & 371 &  381  \\ 
	21    & 878 & 852 & 861 &  890  \\
	22    & 414 & 362 & 418 &  401  \\
	23    & 147 & 162 & 148 &  180  \\
	24-30 & 114 & 87 & 95 &  111  \\
	31-75 & 79 & 78 & 72 &  64 \\
	\hline 
  \end{tabular}
\end{center}
\end{table}
\noindent

\subsection{Implementation Details}

We implemented the IGAE-Net using PyTorch deep learning framework and trained it on NVIDIA GeForce RTX 2080 Ti GPU. The input images are resized to $256 \times 256$ and then randomly cropped to $224 \times 224$, augmented by random horizontal flip, color jittering and normalization during training. However, only normalization is utilized during testing with the test images resized to $224 \times 224$, without a random crop. A random order of images is used by reshuffling the dataset. We use a sum of cross-entropy losses over the predictions of the 3 heads as in Eq.~(\ref{eqn:totalLoss}) to train the IGAE-Net. To prevent over-fitting and over-confidence, label smoothing~\cite{SzeVanIof16} with smoothing value ($\epsilon$) of 0.1 is also used with the cross-entropy loss. We train the model for 50 epochs with mini-batch size of 20 and Adam optimizer with the weight decay factor for L2 regulization of $5 \times 10^{-4}$. For the first 10 epochs, we use a warmup strategy~\cite{LuoGuLia20}, increasing a learning rate linearly from $8 \times 10^{-6}$ to $8 \times 10^{-4}$, and then it is decayed to $4 \times 10^{-4}$ after 30 epochs. The learning rate is divided by 10 for the existing layers of the backbone network i.e. ten times bigger learning rate is given to the newly added layers (FC layers and layer normalization(s) for the 3 heads) with default weight and bias initializations.

\subsection{Performance Evaluation}  

We compare the performance of our proposed IGAE-Net with different up-to-date deep learning architectures as backbone netwowrk. We use both convolution-based and transformer-based deep learning architectures on a publicly available 11k hands dataset~\cite{Mah19} for comprehensive analysis using accuracy evaluation metric. The quantitative performance comparison of the IGAE-Net with different backbone networks is given in Table~\ref{tbl:ComparisonHand}. All the backbone networks are pre-trained on ImageNet-1K~\cite{OlgJiaHao14}.

\textbf{CNN-based Backbone}: For the case of the convolution-based architectures, we use ResNet50~\cite{HeZhaSun16}, DenseNet121~\cite{HuaLiuVan17} and ConvNeXt-Tiny~\cite{ZhuHanCha22}. As shown in Table~\ref{tbl:ComparisonHand}, ConvNeXt-Tiny-based IGAE-Net averagely outperforms all other CNN-based methods across all the sub-datasets in accuracy evaluation metric. 

\textbf{Transformer-based Backbone}: For the case of the transformer-based architectures, we use ViT-B-16~\cite{AleLucAle21}, Swin-T~\cite{ZeYutYue21} and MaxViT-T~\cite{TuTalZha22}. As can be observed in Table~\ref{tbl:ComparisonHand}, the best performing transformer-based IGAE-Net is with the Swin-T backbone network when considering efficiency and accuracy. Though ViT-B-16-based IGAE-Net is performing comparably as the Swin-T-based IGAE-Net, the number of parameters of the ViT-B-16 is almost 3 times that of the Swin-T i.e. 86.6M vs 28.3M (M for million). 

The majority of the models perform better in gender prediction when compared to the identity and age group predictions. For instance, ConvNeXt-Tiny-based IGAE-Net has scored 100\% gender prediction on right dorsal (D-r) of 11k, left dorsal (D-l) of 11k and right palmar (P-r) of 11k hands sub-datasets. 

\begin{figure}[htbp]%[!htb] %[t]%[!h]
  \begin{center}
   \subfloat[] %[Channel Attention Module (CAM)]
  {\label{fig:Identity_cm} \includegraphics[width=0.70\linewidth]{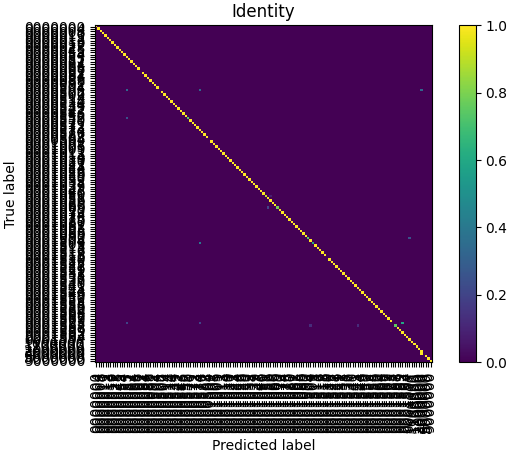}} \\%& %\\ %height=0.46
  \subfloat[] %[Spatial Attention Module with Relative Positional Encodings (SAM-RPE)]
  {\label{fig:Gender_cm} \includegraphics[width=0.70\linewidth]{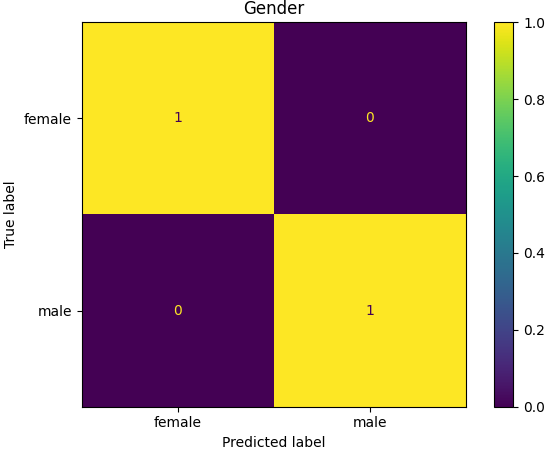}} \\ %& %\\ %height=0.18
    \subfloat[] %[Spatial Attention Module with Relative Positional Encodings (SAM-RPE)]
  {\label{fig:Age_cm} \includegraphics[width=0.70\linewidth]{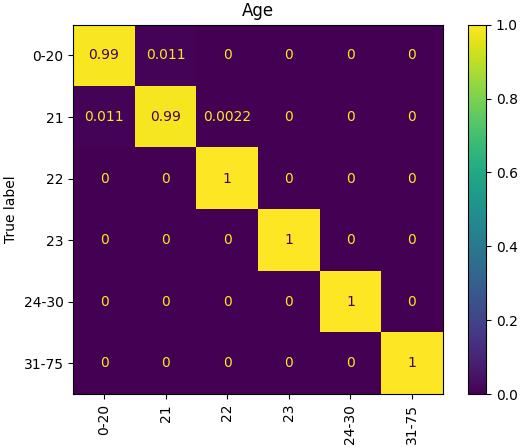}}  %& %\\ %height=0.18
  \end{center}
   \caption{Confusion matrices on right dorsal of 11k hands dataset~\cite{Mah19} using Swin-T~\cite{ZeYutYue21}-based IGAE-Net: (a) confusion matrix for identity outputs, (b) confusion matrix for gender outputs, (c) confusion matrix for outputs of age groups.}
  \label{fig:ConfusionMatrices}
\end{figure}

We also showed the confusion matrices for identity outputs, gender outputs and age groups outputs in Fig. \ref{fig:ConfusionMatrices} on right dorsal of the 11k hands dataset using Swin-T-based IGAE-Net for more detailed analysis of the model's performance. As can be seen on the main diagonal of each confusion matrix, all the cases are predicted as the true label of their class with few exceptions on the identity and age predictions. It is important to note that very few false positive percentage in the age group classification model corresponds to the images belonging to the 21 age group which are mistaken for 0-20 and 22. Similarly, few images belonging to the 0-20 age group are mistaken for 21. The main diagonal of the gender confusion matrix in Fig.~\ref{fig:Gender_cm} confirms the 100\% accuracy of our model on the right dorsal of the 11k hands dataset using Swin-T backbone network given in Table~\ref{tbl:ComparisonHand}.

\begin{figure}[htbp] %[t]%[!htb] %[t]%[!h]
\begin{center}
  \includegraphics[width=1.0\linewidth]{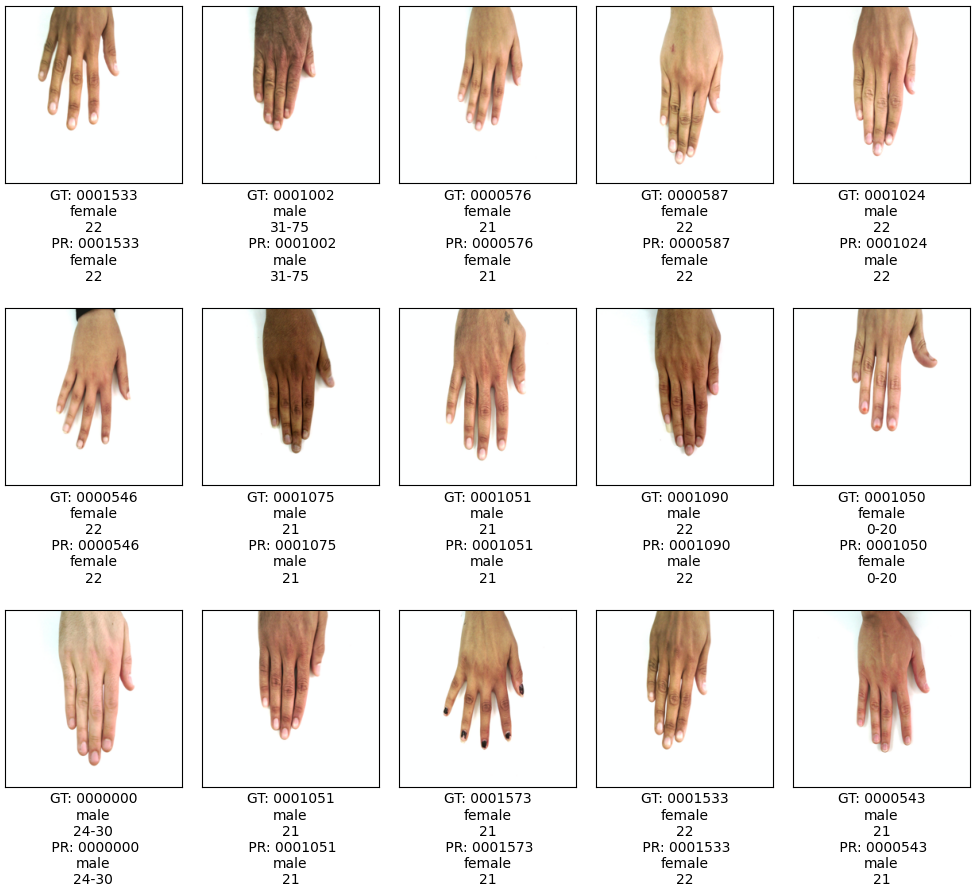} \\%& %\\ %height=0.46
\end{center}
   \caption{Some qualitative results of our proposed method on right dorsal of 11k hands dataset~\cite{Mah19} using Swin-T~\cite{ZeYutYue21}-based IGAE-Net. The ground truth labels (GT) vs the predicted labels (PR) of identity, gender and age group of each hand image, respectively, are shown.}
\label{fig:demoResultH}
\end{figure}

The qualitative results of our proposed method on right dorsal of the 11k hands dataset using Swin-T-based IGAE-Net are also shown in Fig.~\ref{fig:demoResultH}. As can be observed in Fig.~\ref{fig:demoResultH}, the IGAE-Net model predicts correctly for all of  the sample images of our test set in terms of identity, gender and age group. The model’s predictions (PR) are provided next to the ground-truth labels (GT) of the hand images. For instance, for the first hand image in this figure, GT labels are 0001533 (identity), female (gender) and 22 (age group). The model predicted the correct outputs i.e. 0001533 (identity), female (gender) and 22 (age group).

In this evaluation, we assume a closed setting where the training classes are available in the test set with different images from the training set. In open setting such as in real criminal investigation applications in the court, the learned feature representations just after a fully-connected layer and before a softmax function of each head (shown in Fig.~\ref{fig:IGAE-Net}) can be used for feature representations comparison using, for instance, cosine similarity metric.

\begin{table*} [htbp]%[!htb]%[!h]%[tb]
\caption{\normalfont{Quantitative performance comparison of our method (IGAE-Net) with different up-to-date deep learning backbone architectures on right dorsal (D-r) of 11k, left dorsal (D-l) of 11k, right palmar (P-r) of 11k and left palmar (P-l) of 11k hands sub-datasets~\cite{Mah19}. The results are shown in  accuracy (\%) with the best results in $\textbf{bold}$. The models are pre-trained on ImageNet1k~\cite{OlgJiaHao14}.}}
\label{tbl:ComparisonHand}
\begin{center}
  \begin{tabular}{|l|lll|lll|lll|lll|}
    \hline
    \multirow{3}{*}{Method} & 
      \multicolumn{3}{c|}{D-r of 11k} & 
      \multicolumn{3}{c|}{D-l of 11k} & 
      \multicolumn{3}{c|}{P-r of 11k} &
      \multicolumn{3}{c|}{P-l of 11k} \\
      \cline{2-13}
    & Identity & Gender & Age & Identity & Gender & Age & Identity & Gender & Age & Identity & Gender & Age \\
    \hline   
    ResNet50\cite{HeZhaSun16} & 95.82 & 99.78 & 95.39 & 98.83 & 99.46 & 99.63 & 97.45 & 99.00 & 98.82 & 96.61 & 99.27 & 94.71 \\
    %ResNet50-SCG & 97.57 & \textbf{100.00} & 98.32 & 99.05 & 99.58 & 99.37 & 98.65 & 99.72 & 99.47 & 97.99 & 99.12 & 96.22 \\
    DenseNet121\cite{HuaLiuVan17} & 98.75 & 99.61 & 99.12 & 99.61 & \textbf{100.00} & 99.82 & 98.50 & 97.94 & 99.33 & 98.16 & 98.04 & \textbf{98.40} \\
    ConvNeXt-Tiny\cite{ZhuHanCha22} & 98.61 & \textbf{100.00} & 99.78 & \textbf{99.64} & \textbf{100.00} & 99.75 & 98.99 & \textbf{100.00} & 99.55 & \textbf{98.34} & 99.81 & 98.39 \\ 
    \hline
    ViT-B-16\cite{AleLucAle21} & 98.11 & 99.80 & \textbf{99.81} & 99.47 & \textbf{100.00} & 99.67 & \textbf{99.47} & \textbf{100.00} & \textbf{99.67} & 98.17 & 99.12 & 97.36 \\ 
    Swin-T\cite{ZeYutYue21} & 98.08 & \textbf{100.00} & 99.54 & 99.47 & \textbf{100.00} & \textbf{99.89} & 99.10 & 99.92 & \textbf{99.67} & 97.18 & \textbf{99.84} & 97.66 \\
    MaxViT-T\cite{TuTalZha22} & \textbf{99.04} & \textbf{100.00} & 98.95 & 99.40 & 99.37 & 98.80 & 98.21 & 99.80 & 98.94 &  96.13 & 99.35 & 93.19 \\
	
	\hline 
  \end{tabular}
\end{center}
%\caption{Quantitative performance comparison of our method (LAGA-Net) with other methods (GPA-Net~\cite{BaiWilHosGPA21}, MBA-Net~\cite{BaiWilHosMBA21}, RGA-Net~\cite{ZhiCuiWen20} and ABD-Net~\cite{TiaShaJin19}) on right dorsal (D-r) of 11k, left dorsal (D-l) of 11k, right palmar (P-r) of 11k, left palmar (P-l) of 11k and HD datasets. The results are shown in rank-1  accuracy (\%) and mAP (\%). Best and second best results are shown in $\color{red}{\textbf{red}}$ and $\color{blue}{\textbf{blue}}$, respectively.}
%\label{tbl:ComparisonHand}
\end{table*}
\noindent

\noindent

\noindent

\section{Conclusion}  \label{sec:Conclusion}

In this work, we introduce a multi-task representation learning framework to jointly estimate the identity, gender and age of individuals from their hand images for the purpose of criminal investigations. We investigate our proposed identity, gender and age estimation network (IGAE-Net) with different up-to-date deep learning architectures as backbone network and compare their performance for joint estimation of identity, gender and age from hand images of perpetrators of serious crime such as sexual abuse. To simplify the age prediction, we create age groups for the age estimation. We make extensive evaluations and comparisons of both convolution-based and transformer-based deep learning architectures on a publicly available 11k hands dataset, which demonstrates that it is possible to efficiently estimate the identity as well as other attributes such as gender and age of suspects jointly from their hand images for criminal investigations. From the convolution-based architectures, ConvNeXt-Tiny-based IGAE-Net performs best whereas Swin-T-based IGAE-Net is the best performing transformer-based architecture. These two networks, ConvNeXt-Tiny-based IGAE-Net and Swin-T-based IGAE-Net, give comparable performance, which confirms the competitive nature of the convolution-based and transformer-based architectures for computer vision tasks. Our proposed method is crucial in assisting international police forces in the court to identify and convict abusers.

%\section*{Acknowledgment}
%We would like to thank Bryan Williams, Hossein Rahmani, Plamen Angelov and Sue Black who were part of the project which our preliminary work~\cite{BaiWilHosMBA21} is associated to.


\bibliographystyle{IEEEbib}
\bibliography{refs}

\vfill

\end{document}